\documentclass[10pt,twocolumn,letterpaper]{article}

\usepackage{iccv}
\usepackage{times}
\usepackage{epsfig}
\usepackage{graphicx}
\usepackage{amsmath}
\usepackage{amssymb}

\usepackage[breaklinks=true,bookmarks=false]{hyperref}

\iccvfinalcopy % *** Uncomment this line for the final submission

\def\iccvPaperID{****} % *** Enter the ICCV Paper ID here

\begin{document}

%%%%%%%%% TITLE
\title{MSC: Multi-Scale spatio-temporal Causal attention \\for Autoregressive Video Diffusion}

\author{Xunnong Xu \hspace{20pt} Mengying Cao\\
{\tt\small xunnongxu@gmail.com \hspace{20pt} mengyingcao@foxmail.com}
% For a paper whose authors are all at the same institution,
% omit the following lines up until the closing ``}''.
% Additional authors and addresses can be added with ``\and'',
% just like the second author.
% To save space, use either the email address or home page, not both
% \and
% Second Author\
% \and 
% Third Author\
% Institution2\\
% First line of institution2 address\\
% {\tt\small secondauthor@i2.org}
}

\maketitle
%\thispagestyle{empty}

%%%%%%%%% ABSTRACT
\begin{abstract}
   Diffusion transformers enable flexible generative modeling for video. However, it is still technically challenging and computationally expensive to generate high-resolution videos with rich semantics and complex motion. Similar to languages, video data are also auto-regressive by nature, so it is counter-intuitive to use attention mechanism with bi-directional dependency in the model. Here we propose a Multi-Scale Causal (MSC) framework to address these problems. Specifically, we introduce multiple resolutions in the spatial dimension and high-low frequencies in the temporal dimension to realize efficient attention calculation. Furthermore, attention blocks on multiple scales are combined in a controlled way to allow causal conditioning on noisy image frames for diffusion training, based on the idea that noise destroys information at different rates on different resolutions. We theoretically show that our approach can greatly reduce the computational complexity and enhance the efficiency of training. The causal attention diffusion framework can also be used for auto-regressive long video generation, without violating the natural order of frame sequences. 
\end{abstract}

%%%%%%%%% BODY TEXT
\section{Introduction}
Diffusion models \cite{sohl2015deep, ho2020denoising} make it possible to generate highly realistic and detailed images \cite{rombach2022high}. With the extra frame dimension, video generation is the next major milestone for generative modeling. The DiT model \cite{peebles2023scalable}, where the diffusion framework is combined with the flexible transformer architecture, became the standard approach to video generation \cite{openaisora2024, yang2024cogvideox, polyak2024movie}.  

In most DiTs, there is only one spatial scale, which is represented in the patchify step after compressing video into latents, and before going through subsequent transformer layers for diffusion. For example, a video clip of shape $3 \times T \times H \times W $ becomes $C\times (T/8q) \times (H/8p) \times (W/8p)$ after compression and patchification, where both the image and time compression factors are 8, while $q\times p \times p$ is the spatio-temporal patch size \cite{polyak2024movie}. When the image resolution is high or when the video is long, the total number of tokens can be very large, which leads to large computational complexity and GPU memory consumption. Another problem with the single scale approach is that real-world videos have different levels of spatial details in individual image frames, as well as different patterns of motion happening at different time scales. Using multiple spatio-temporal scales have two benefits:  reduced computation complexity for attention calculation, and improved efficiency for the model to learn the rich semantics and complex motion patterns in the data.

Video data are also auto-regressive in the frame (or time) dimension, but this has mostly been ignored by mainstream video generation models. Existing approaches usually treat image patches as tokens and then flatten them into a long 1-D sequence, as in language models. There are two ways to model the relationship between tokens. The non-autoregressive approach (a more popular one) uses bi-directional attention, while the auto-regressive approach uses masked causal attention, both of which are at the token/patch level but not at the frame level. Here we introduce a frame-level causal attention to better model the nature of video data. Furthermore, when combining frame-level causal attention with the diffusion framework, the partial masking problem when conditioning on noisy frames arises. We reuse the idea of multiple scales to alleviate this, considering that image features at different spatial resolutions are affected differently by the added noise in the forward diffusion process. 

Our main contributions are summarized as follows:
\begin{itemize}
    \item We proposed a multi-scale spatio-temporal causal attention framework for auto-regressive video diffusion generation. 
    \item The multi-scale learning combined with local sliding-window attention and strided global attention in both the spatial and temporal dimensions can substantially reduce the computational complexity of transformer based video diffusion. 
    \item We design frame-wise causal attention based on multiple scales, so causal conditioning on noisy image frames with independent noise levels can still be effective in diffusion training.  
\end{itemize}

%------------------------------------------------------------------------
\section{Related Work}
\noindent\textbf{Video generation models.} There are a lot of video generation models works by combining image generation backbone such as pretrained Unet \cite{rombach2022high} with additional temporal layers for frame consistency \cite{blattmann2023align}. With the introduction of Sora \cite{openaisora2024}, many new models with similar architectures and capabilities are being introduced, including Open-Sora \cite{opensora}, Open-Sora-Plan\cite{lin2024open}, CogVideoX \cite{yang2024cogvideox}, Meta Movie-Gen \cite{polyak2024movie}, Runway Gen-3 alpha \cite{gen3alpha2024}. Most of them use the Diffusion Transformer architecture on spatially and temporally compressed videos for diffusion training, with different considerations for applying temporal consistency. 

\noindent\textbf{Multi-scale learning.} The idea of multi-scale learning is not new to computer vision. One branch of study along the line is to use hierarchical structures for image classification and object detection, for example from the AlexNet \cite{krizhevsky2017imagenet} to the UNet\cite{ronneberger2015u} and the feature pyramid networks \cite{lin2017feature}. In these models, there is usually a sequence of layers operating on decreasing spatial resolutions, moving from detecting high-resolution low-level features in shallow layers to detecting low-resolution high-level features in deep layers. The other branch of study is to use blocks with different resolutions in parallel rather than in sequence, such as in Inception network \cite{szegedy2015going}, InceptionFormer \cite{si2022inception}, LITv2 \cite{pan2022fast}, Swin Transformer \cite{liu2021swin}, with the goal of improving the efficiency of networks. Inside a single layer, there are parallel blocks operating on different spatial resolutions, and their outputs are concatenated along the feature dimension before sending to the next layer. We borrow the ideas from LITv2 \cite{pan2022fast} with parallel branches but generalize it to multiple resolutions instead of two and use varying resolutions for different layers. We also consider the temporal dimension of the data in addition to the spatial dimension. 

\noindent\textbf{Causal or Autoregressive models.} There is a lot of study on GPT-like auto-regressive structure for image generation, such as ImageGPT\cite{chen2020generative} for next token prediction or VAR \cite{tian2024visual} for next-scale prediction. They become important alternatives to diffusion models. It it however very challenging to directly combine the diffusion framework with auto-regressive structures, because in diffusion tokens are usually denoised in parallel with bi-directional but not causal dependency. There are some attempts in this direction, for example in DiffusionForcing \cite{chen2024diffusion} and AR-Diffusion \cite{wu2023ar}. In these models, tokens at different positions have different noise levels, and they implemented special techniques to address the the partial masking problem for conditioning on noisy tokens. We advance these ideas further by integrating them with multiple resolutions of image features. 

%------------------------------------------------------------------------
\section{Methods}
We propose a multi-scale spatio-temporal causal diffusion transformer (MSC) as a foundational architecture for video generation. This model is composed of a stack of $L$ MSC transformer layers, which is similar to the regular transformer but with the self-attention block replaced by a multi-scale attention block. This multi-scale block is divided into several resolution branches in parallel, and the attention outputs from them are concatenated (with optional up-sampling operation) before going to the feed-forward layer. To better model the relationship between image frames in video, we use causal attention in instead of bi-directional attention in all resolution branches. The diffusion timestep or noise level embedding is used to control the relative weights of different resolution branches, such that information on multiple resolutions can flow efficiently from past frames to future frames even when independent noise are added to them in diffusion training.   

%------------------------------------------------------------------------
\subsection{Multi-Resolution spatial learning}
\label{sec31}

\begin{figure*}[!t]
\begin{center}
\includegraphics[width=1\linewidth]{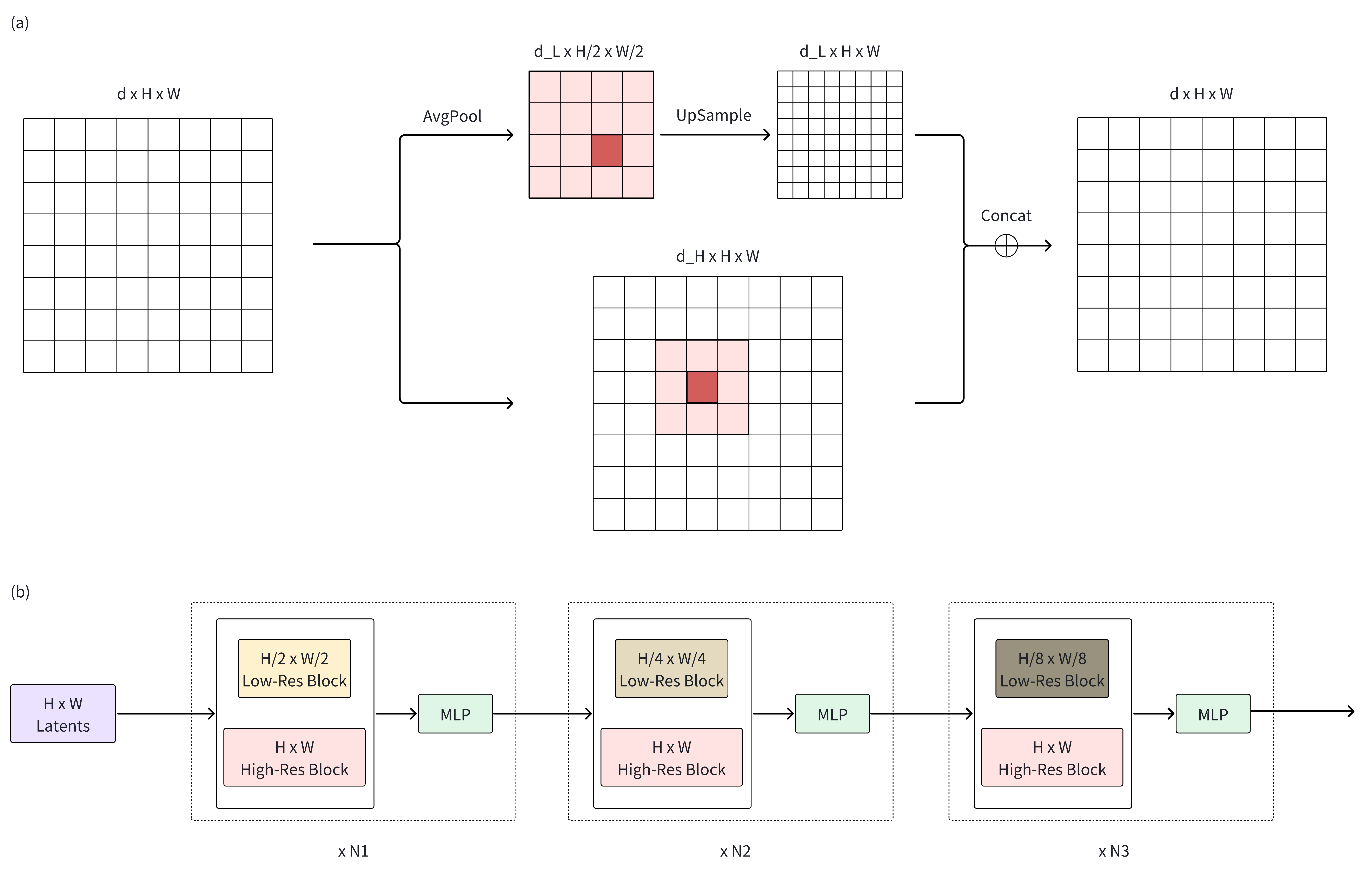}
\end{center}
   \caption{Multi-Resolution spatial learning framework: (a) two spatial branches in each transformer layer, with local sliding window attention for High-Res branch and global attention for Low-Res branch; (b) more down-sample in the Low-Res branch of deeper transformer layers.}
\label{fig1}
\end{figure*}

Most of the state-of-the-art text to image or text to video generation models use the Diffusion Transformer (DiT) architecture as backbone. In this approach, image or video data are represented as patches with fixed size in both the spatial and temporal dimension. These patches are equivalent to the tokens in large language models. However, real-world video data often capture rich information at different spatial resolutions and time scales. A widely used architecture for text to image generation before DiT is the U-Net \cite{ronneberger2015u}, which encodes information at increasing spatial scales via a series of down-sampling layers, and then recombines those information when up-sampling all the way to original image resolution. With a low spatial resolution (or larger patch size), we can have the overall style and setting of a scene. At high resolution (or smaller patch size), there are a lot of details for static contents and complex motions for dynamic objects. Finally, using multiple spatial resolutions make it easier to generalize to high resolution data, because the context length for attention scales quadratically with image resolution, and the overall computational complexity of attention calculation is another quadratic dependency on the context length.

We want to combine the efficiency of multi-resolution modeling with the flexibility of the transformer architecture, and design a multi-resolution transformer for video modeling, where both efficiency and flexibility are necessary. Our proposed transformer has two spatial branches, a High-Res branch that operates at a default high resolution, and a Low-Res brach that operates at a lower resolution with down-sampling operations over the spatial dimension. The original multi-head attention in each transformer layer is divided into these resolution branches along the head dimension. Since high resolution features encode more high frequency details spatially and low resolution features encode more low frequency information, we can then set different attention windows for these branches, as in \cite{pan2022fast}. Specifically, the High-Res branch has a limited attention window, which should be enough for encoding local high frequency information.  The Low-Res attention has a larger attention window, ideally covering the whole image frame, such that the semantics and high-level global features will not be lost. The sliding window attention mechanism can be easily implemented with the Neighborhood Attention package \cite{hassani2023neighborhood}. Finally, analysis \cite{dosovitskiy2021imageworth16x16words} shows that shallow layers tend to learn high resolution but low-level features, while deep layers tend to learn low resolution but high-level features. We can design our network accordingly, for example, choosing a larger down-sample factor for the low-res branch in deeper transformer layers. These three aspects of our multi-scale spatial learning architecture illustrated in Fig.(\ref{fig1}).

We also discuss why choosing parallel branches rather than sequential or hierarchical structures for multi-scale learning. The arguments are as follows. Hierarchical structures or feature pyramids are good for classification and object detection tasks. However, here we are doing per-pixel denoising, so it might not be the best idea to use these sequential structures. For parallel branches,  we can mix the features from multiple resolutions in every transformer layer, as suggested in Inception Transformer \cite{si2022inception}, which leads to between information exchange between resolutions. Finally, since the $L$ transformer layers are almost identical in structure, it is easier to implement pipeline parallelism for efficient training. This design is also conceptually similar to MarDini \cite{liu2024mardinimaskedautoregressivediffusion} model for video generation, where they have high-res and low-res models running in parallel and with information exchange in certain stages.

%------------------------------------------------------------------------
\subsection{Hi-Lo frequency temporal learning}
\label{sec32}
Similar to the spatial dimension, real-world video data also contain rich information at different time scales. The motion patterns in video can be roughly recognized as two categories, the global motion of the whole scene due to camera movement or rotation, and the local motion of dynamic objects that occupies part of the scene. For every motion pattern, either global or local, we can associate a speed of movement with it. This means that image patches in nearby frames can only affect image patches that are also nearby spatially, otherwise it will imply that the speed of movement of dynamic objects is too high, which is not physically allowed. In other words, the effective attention window size should be roughly proportional to the frame difference between the query and key patches/tokens. This is similar to the idea of light-cone in physics, which sets a boundary in space-time for establishing causal relationship between events. 

\begin{figure}[t]
\begin{center}
\includegraphics[width=1\linewidth]{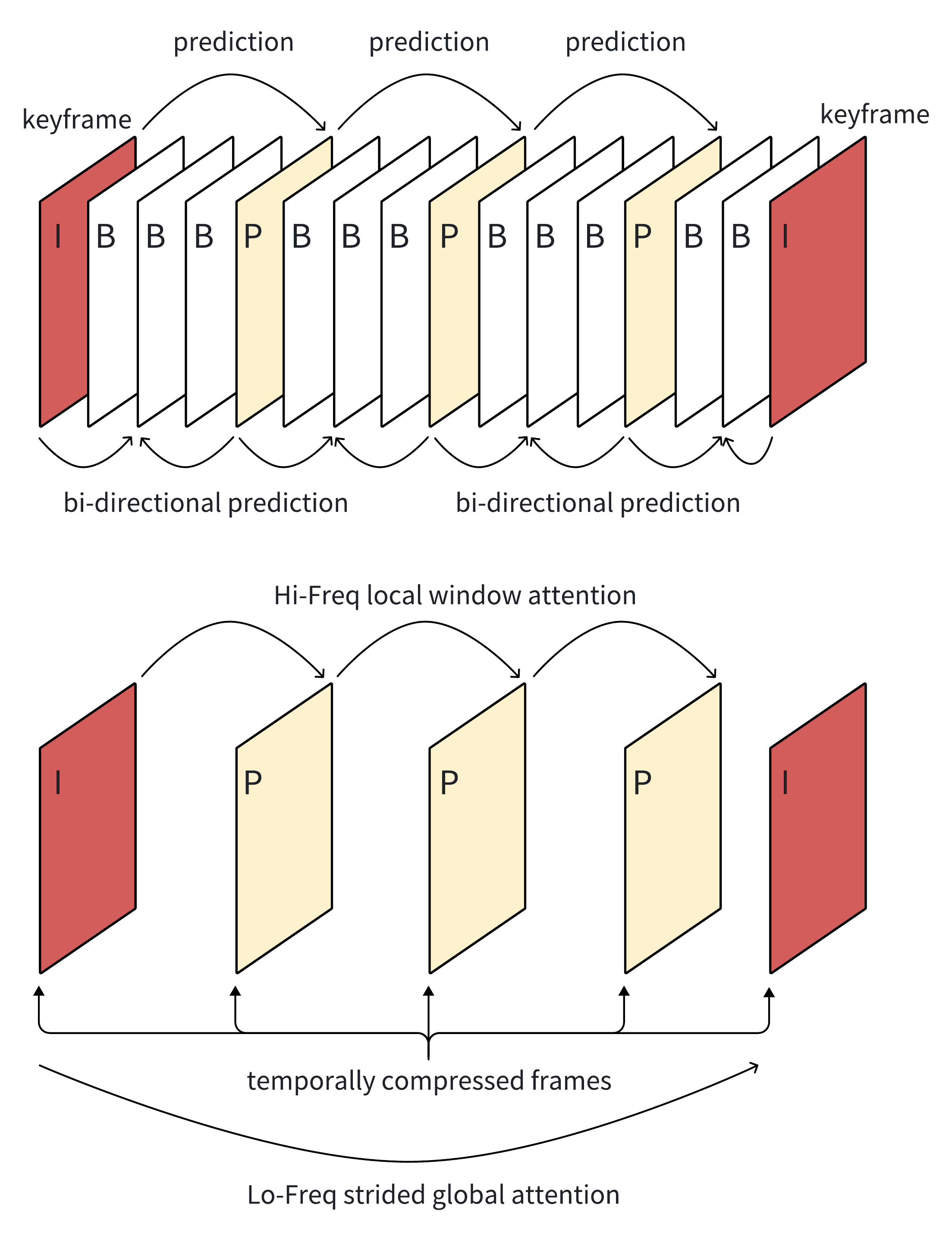}
\end{center}
   \caption{After temporal compression with 3D-VAE, the IPB frame structure of video data clearly show two time scales: the low frequency I frames with long-range dependency and the high frequency P frames with short-range dependency.}
\label{fig2}
% \label{fig:onecol}
\end{figure}

Video data have a lot of redundancy between frames, which is the subject of study of many inter-frame compression algorithms. In technical terms \cite{beach2018video}, there are key frames ($I$-frame), prediction frames ($P$-frame) and bi-directional prediction frames ($B$-frame) between key frames in compressed video. The distance between $I$-frames or group-of-picture (GOP) can be quite large (e.g. $>100$), depending on the data content and compression settings. By definition, $P$-frames are predicted from a most recent $I$-frame or $P$-frame, and $B$-frames are predicted from two adjacent $I$-frame or $P$-frame. For example, in a GOP of the form $IBBBPBBBPBBBPBBI$ as in Fig.(\ref{fig2}), there are 3 $P$-frames predicted sequentially, and in-between $I-P$ or $P-P$ frames there are several $B$-frames predicted bi-directionally. We argue that the common technique of temporal compression in the 3D VAE \cite{yu2023language, yang2024cogvideox, polyak2024movie} of many recent video generation models, is conceptually equivalent to the bi-directional prediction $B$-frame in video compression. There are three things in common for these two concepts: 1) The compression happens in the finest granularity, which is at the frame level; 2) the dependence between frames is bi-directional, so there is no need for causal modeling; 3) the factor of temporal compression is usually 4 or 8, which roughly equals the number of $B$-frames in-between $P$-frames. With temporal compression, we can ignore the concept of $B$-frames completely and only focus on $I$-frames and $P$-frames. In our example we are left with a new frame sequence $IPPPI$, which clearly show two time-scales, as shown in Fig.(\ref{fig2}). At the coarser time scale, we have the $I$-frames, where the distance between adjacent frames is the GOP, and the actual time difference is in the order of seconds. At the finer time scale, we have the $P$-frames, where the relationship between frames is strictly local, and the actual time between frames is in the order of 100 milliseconds. 

We design attention blocks with appropriate temporal attention window and stride to model the dual time-scale structure discussed above. Specifically, for the High-Res spatial branch where high-frequency and local details are being modeled, we choose a finite temporal attention window in the frame dimension, similar to the $P$-frames. An image patch in frame $t$ will only attend to $w$ previous frames, which can realized with sliding window neighborhood attention mechanism \cite{hassani2023neighborhood}. For the Low-Res spatial branch, where frame-global low frequency information is considered, attention can be calculated between more distant frames but at a much coarser granularity, just like the $I$-frames. One way to model this is to use dilated or strided attention \cite{hassani2022dilated} over the frame axis. 
\begin{figure}[!th]
\begin{center}
\includegraphics[width=1\linewidth]{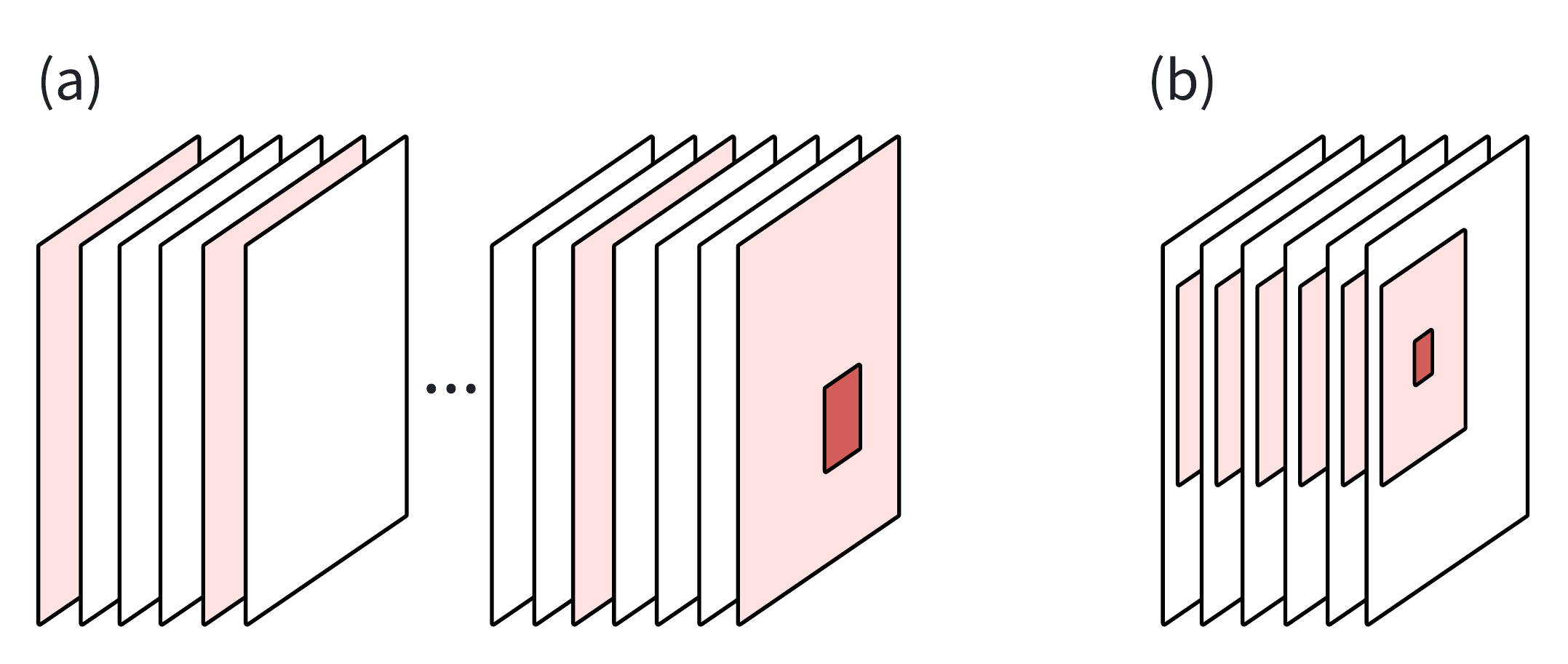}
\end{center}
   \caption{Hi-Lo frequency temporal learning framework: (a) strided global  attention for the Low-Res branch. (b) local sliding window attention for the High-Res branch. In both figures, the dark-red image patch is the attention query, while the light-red image patches are the attention key and value.}
\label{fig3}
% \label{fig:onecol}
\end{figure}

We now discuss the computational complexity of the proposed multi-scale spatio-temporal learning framework. Compared to vanilla diffusion transformer, most of the changes are in the attention block, while the feed-forward layer stay the same, so the reduction in computational complexity comes from the improved efficiency of the new MSC attention mechanism. For regular attention with batch size $b$, sequence length $s$ and hidden dimension $h$, the complexity is $6bsh^2 + 2bs^2h + 2bsh^2 = 8bsh^2 + 2bs^2h$, which includes linear projection, self-attention, and out projection. We consider the case when the hidden dimension is divided equally between the two parallel branches, with $h_L = h_H = h/2$. For the High-Res branch, we have a much smaller sequence length due to the local spatial-temporal windows, so the complexity is $6bs(h/2)^2 + 2bs^2(h/2)^2/(w^2v) + 2bs(h/2)^2 = 2bsh^2 + bs^2h/(2w^2v)$, where $w$ and $v$ are the spatial and temporal attention window. For the Low-Res branch, the down-sampling leads to reduction in total number of tokens, as well as reduced complexity in the linear projections of $QKV$. Furthermore, each frame attends to other frame in a strided fashion, so we have another reduction factor $d$. The complexity is $6b(s/r^2)(h/2)^2 + 2b(s/r^2)(s/r^2/d)(h/2)^2 + 2b(s/r^2)(h/2)^2 = 2bsh^2/r^2 + bs^2h/(2r^2d)$, with $r$ as the down factor. The total theoretical complexity is greatly reduced with contributions from the separation of scales, down-sample, local attention and strided attention. 

%% maybe we should put a table here?

%------------------------------------------------------------------------
\subsection{Noise scale modulated causal attention}
\label{sec33}
Video data, which consist of sequence of image frames, are intrinsically auto-regressive and thus very similar to languages. However, this similarity only lies in the equivalence between tokens in text and individual image frames in a video, but not in the image patch level. Most DiT models treat image patches as tokens and uses GPT-like network \cite{dubey2024llama} to model the relationship between tokens, without even considering the causal dependency. This approach does not correctly characterize the sequential nature of video data. Moreover, bi-direction attention between tokens makes it difficult to generate long videos or realize interactive control of video content, just like the difference between causal decoder-only GPT \cite{radford2018improving} and bi-directional language transformers such as BERT \cite{devlin2018bert}. 

As mentioned in Sec.\ref{sec32}, after temporal compression with 3D VAE, there is no need to model the $B$-frames and we are left with a sequence of $I$ and $P$ frames $IPPPI$. By definition, the $P$-frames are predicted from previous $I$ or $P$ frames, so its modeling is strictly causal. For the $I$-frames, they are independent key frames and can be encoded with image compression only. Their causal dependency comes from the physical nature of the scene happening in real-world: time flows only in the forward direction and an event can only affect future events but not past ones. With these analysis in mind, it is necessary to have causal dependency built-in to our attention mechanism. To implement causal attention, we need to make sure that an image patch or token only attend to tokens in previous frames: 
$ \mathrm{CausalAttention} = f(x_{tij}, x_{t^{\prime}kl})$,  where $t^{\prime} \leq t$.

\begin{figure}[t]
\begin{center}
\includegraphics[width=1\linewidth]{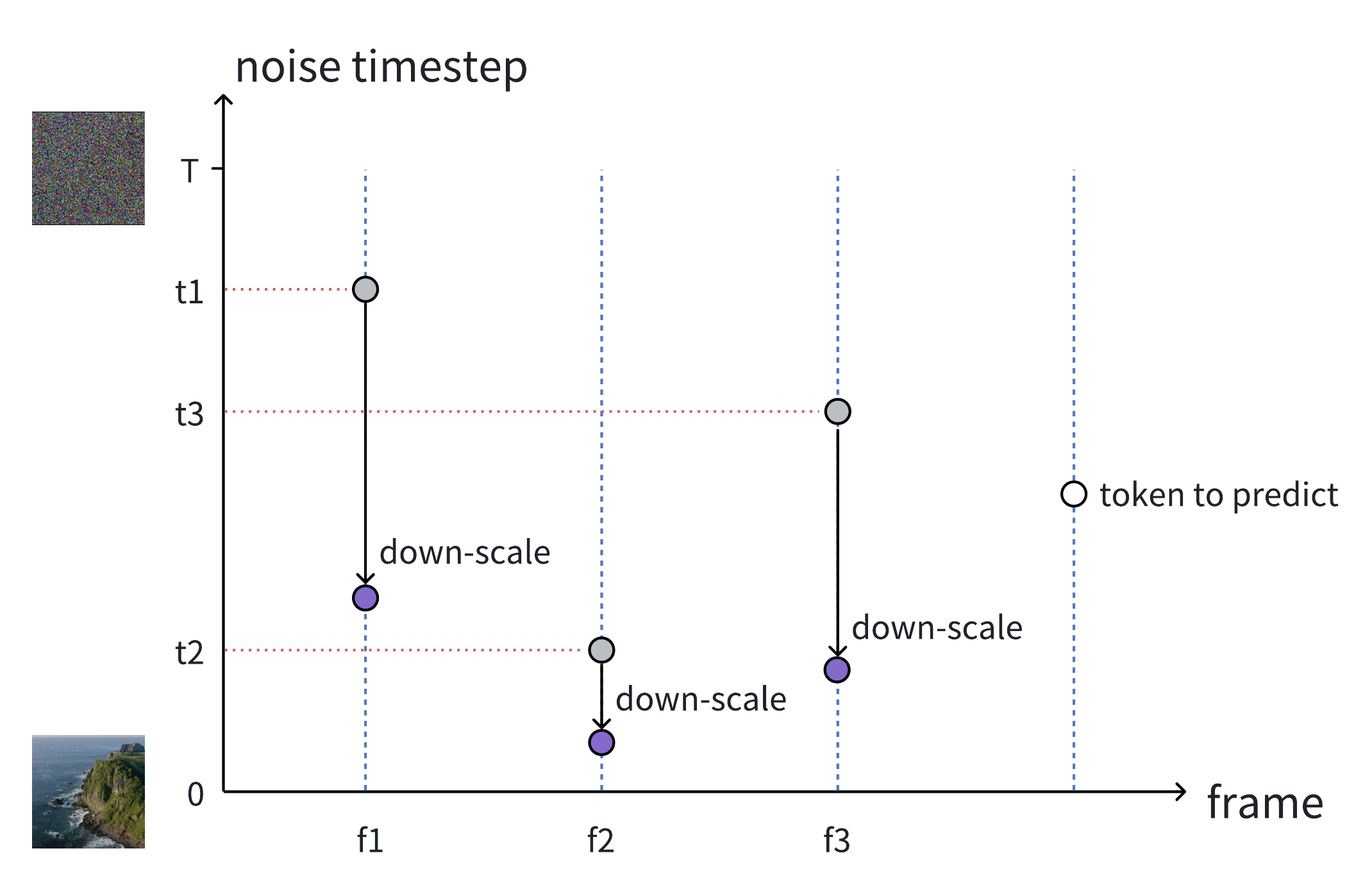}
\end{center}
   \caption{In our proposed frame-wise causal attention, each token-to-be-predicted can only be affected by tokens in previous frames, but not by those in future frames. During diffusion training, we use independent noise timesteps for tokens in different frames. When applying down-sampling operations, image features becomes less noisy, which is equivalent to using a smaller noise timestep for diffusion. In this way, conditioning on noisy frames in causal attention becomes more effective.}
\label{fig4}
\end{figure}

Conceptually, we are treating frames as the basic element for auto-regressive modeling. As a result, in diffusion training, we must use independent time steps or noise levels for different frames. This idea is similar to those proposed in AR-Diffusion \cite{wu2023ar} and DiffusionFrocing \cite{chen2024diffusion}. Problem arises when applying causal conditioning on these noisy frames for auto-regressive generation. This main issue is that when a frame gets noisy in the forward diffusion process, conditioning on it for causal attention becomes ambiguous. Noisy frames could be quite different from clean frames in the original video, which makes it harder to train a conditional model, since some of the information is destroyed when adding noise. One way to get around this is to use an increasing noise schedule in the frame dimension, for example adding less noise to older frames and more noise to later frames, as in AR-Diffusion. The intuition is that we always condition on previous frames with less noise and more information. This technique could alleviate the problem to some extent, but it lacks the flexibility of using independent noise levels on different frames for diffusion training. 

Our observation is that, the noise added in the forward diffusion process destroys image features of different spatial resolutions at different rates. Specifically, when the noise level is low, the noisy image still has a lot of details. When the noise level is high, most of the details are missing, and we only get a rough idea of what the image looks like. It basically means that high resolution features are destroyed much more quickly than low resolution features when increasing noise, due to the high-frequency nature of Gaussian noise. In other words, low resolution features becomes more relevant compared to high resolution features at higher noise levels. This is in accordance with the results of \cite{chen2023importance}, where they study the importance of noise scheduling for diffusion models. From a mathematical point of view, the noise at different pixel locations are independent Gaussians, but the original pixel values are usually strongly correlated, especially for nearby pixels. After down-sampling operations such as average pooling, the strength of the signal stays almost the same, but the variance of noise is greatly reduced, which leads to higher signal-to-noise ratio (SNR). For example, with a down factor of 2 in each spatial dimension, the variance of the average noise is reduced by a factor of 4: $\mathrm{AvgNoise} =  \frac{1}{4}\sum_{i=1}^{4} N(0, \sigma_i^2) \sim \frac{1}{4} N(0, 4\sigma^2) \sim  N (0, \sigma^2/4)$ when $\sigma_i^2 = \sigma^2$. By down-sampling, we thus get smaller images with much less noise. These observations are described in more details in Fig.(\ref{fig4}).

\begin{figure}[t]
\begin{center}
\includegraphics[width=1\linewidth]{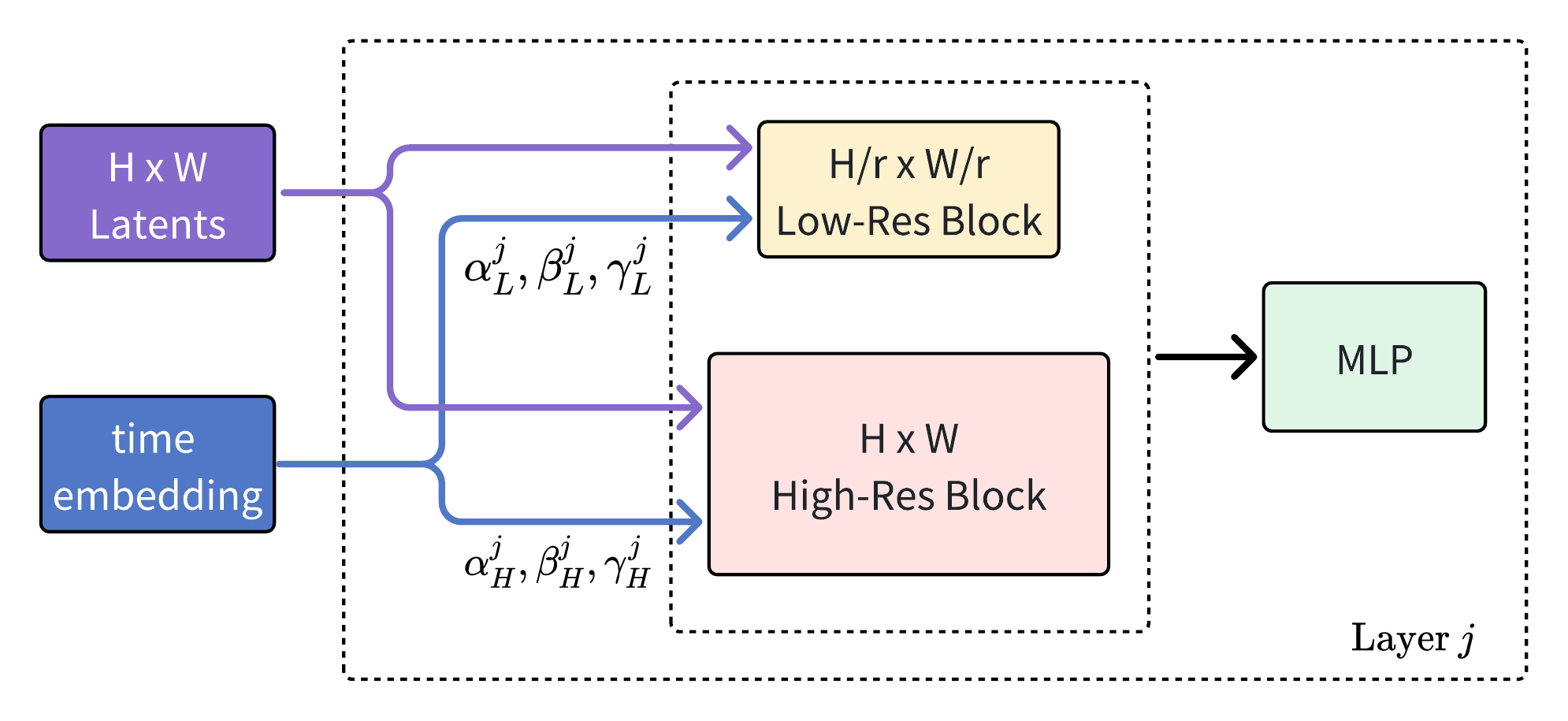}
\end{center}
   \caption{In each transformer layer, the timestep embedding is used to control the weights of the two parallel spatio-temporal branches. By stacking a series of such layers with different down-factor $r$ and with independently learned weights, the full network is able to pass information efficiently for noisy frames during causal conditioning.}
\label{fig5}
% \label{fig:onecol}
\end{figure}

Based on these analysis, we can use the noise level to control the weights of different resolution components, and the concept of conditioning on a noisy image is thus still well-defined for each resolution. To model the noise level dependent weights, we can pass the diffusion timestep or noise level as embedding to the different resolution branches in a transformer layer. The model can then learn the corresponding weights appropriately. To this end, the causal dependence between frames is modeled using independent per-frame noise level and causal attention, as shown in Fig.(\ref{fig5}). This new type of attention is also being seamlessly integrated with the multi-resolution learning framework proposed in Sec.\ref{sec31}.  

% \section{Experiments}
% \subsection{Experiment on the mixkit dataset?}

% \subsection{Experiment on the ByteDance sim dataset?}

\section{Conclusion}
We proposed a multi-scale space-time causal attention framework (MSC) for efficient auto-regressive video diffusion. We separate the single-scale transformer into high resolution and low resolution scales in the spatial dimension, and then combined them with the high-low frequency decomposition in the temporal dimension. These new mechanisms are implemented with local sliding window attention and strided global attention spatio-temporally for effective attention calculation of video data. The idea of multi-resolution spatial learning is further integrated into the causal auto-regressive diffusion framework to achieve reliable information passing between frames at different resolutions. It is worth noting that our MSC is a general framework for video modeling, so it applies to pixel space diffusion models as well as to latent space diffusion models. We leave the detailed investigation of the effectiveness of different design choices to future work.  

{\small
\bibliographystyle{ieee}
\bibliography{egbib}
}

\end{document}